\documentclass[conference]{IEEEtran}
\IEEEoverridecommandlockouts
\usepackage{cite}
\usepackage{amsmath,amssymb,amsfonts}
\usepackage{algorithmic}
\usepackage{graphicx}
\usepackage{booktabs}
\usepackage{textcomp}
\usepackage{multirow}
\usepackage{graphicx}
\usepackage{xcolor}
\usepackage{amsmath}
\usepackage{array}
\usepackage{threeparttable}
\usepackage{makecell}
\usepackage{overpic}
\def\BibTeX{{\rm B\kern-.05em{\sc i\kern-.025em b}\kern-.08em
    T\kern-.1667em\lower.7ex\hbox{E}\kern-.125emX}}
\begin{document}

\title{InstantStyleGaussian: Efficient Art Style Transfer with 3D Gaussian Splatting\\
}

\author{
		Xin-Yi Yu ,Jun-Xin Yu ,Li-Bo Zhou ,Yan Wei and Lin-Lin Ou
}

\maketitle

\begin{abstract}
We present InstantStyleGaussian, an innovative 3D style transfer method based on the 3D Gaussian Splatting (3DGS) scene representation. By inputting a target-style image, it quickly generates new 3D GS scenes. Our method operates on pre-reconstructed GS scenes, combining diffusion models with an improved iterative dataset update strategy. It utilizes diffusion models to generate target style images, adds these new images to the training dataset, and uses this dataset to iteratively update and optimize the GS scenes, significantly accelerating the style editing process while ensuring the quality of the generated scenes. Extensive experimental results demonstrate that our method ensures high-quality stylized scenes while offering significant advantages in style transfer speed and consistency.
\end{abstract}


\section{Introduction}
With the rapid development of applications such as robotics simulation, virtual reality, and autonomous driving, the editing of 3D scenes and models is playing an increasingly important role. Developing user-friendly 3D representations and editing algorithms has always been a key objective in the ever-evolving field of computer vision. Traditional representations such as meshes and point clouds have been widely used for their interactivity in editing, but they still face challenges in editing complex scenes and fine details. Rapidly editable 3D representations enable artists, developers, and researchers to generate valuable content quickly.

Recently, with the advent of implicit neural reconstruction methods like NeRF, capturing real-world 3D scenes has become simple and fast. By obtaining a set of scene images and corresponding camera parameters, these images can be used to continuously optimize the Neural Radiance Field. Compared to traditional representations, NeRF's process is more mature and easier to use, but its implicit nature makes it less intuitive for editing than conventional methods. Nevertheless, it has garnered significant attention, with many extensions derived from its framework for 3D editing as a foundational method \cite{ref1,ref2,ref3,ref4}. However, the lengthy training and rendering times for scene reconstruction hinder its practical application. Although there is research \cite{ref5,ref6} attempting to accelerate this process, it still struggles to meet real-time requirements.

\begin{figure*}[htbp]
\centering
\includegraphics[width=5.5in]{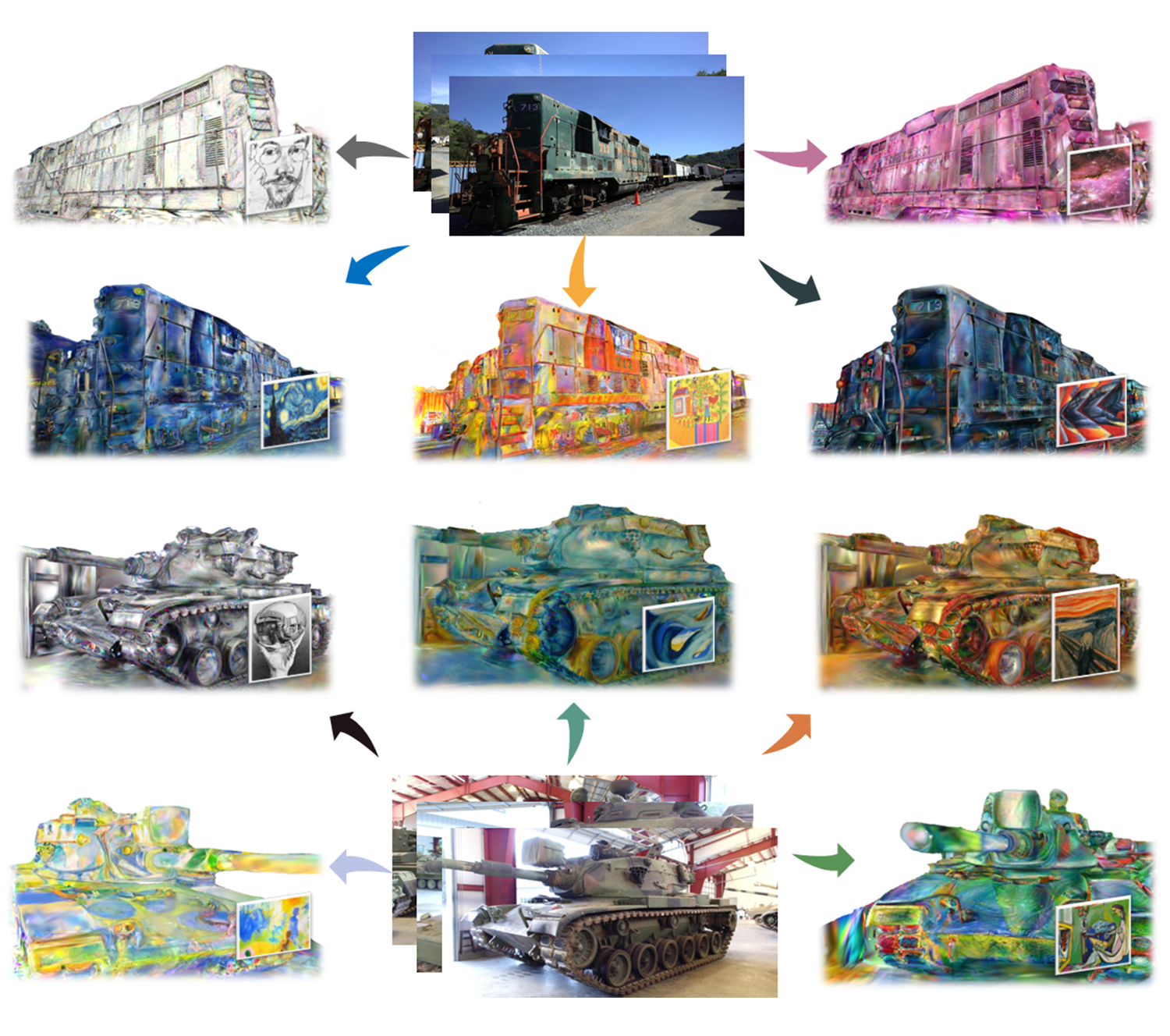}
\caption{We introduce InstantStyleGaussian, an innovative 3D style transfer pipeline. By inputting a target style image, editing can commence, enabling swift style transformation while ensuring consistency across multiple views. The experimental section demonstrates additional scenes from new perspectives following the stylization process.}
\label{fig_1}
\end{figure*}

To create valuable 3D content, performing more diverse edits on 3D scene representations, such as editing surface color and texture features, and achieving 3D style transfer is necessary. Instant interactivity is crucial for a good 3D user experience, requiring editing without sacrificing real-time rendering and multi-view consistency. Recently, 3D Gaussian Splatting (3DGS) \cite{ref7} has enabled fast rendering, and its explicit representation allows for real-time 3D style transfer. Traditional transfer methods require extracting VGG features \cite{ref8} from a style image, embedding them into the reconstructed 3D scene, and decoding these features to render the new scene. However, this process often demands extensive memory and computation time. The decoding method influences the final style transfer effect \cite{ref9,ref10}, which can compromise multi-view consistency and degrade the overall scene quality.

We have designed a 3D style transfer pipeline that allows for the rapid editing of existing 3DGS scenes using only a target style image. This approach enables the transfer of 2D image styles to 3D scenes while maintaining multi-view consistency. Our approach captures images from the camera perspectives of the 3DGS scene and uses a diffusion model to edit multiple perspective images. Based on the Iterative Dataset Update (IDU) method, we propagate the newly generated artistic-style scene images back to create a new 3D scene iteratively. Our method is image-based, which is more intuitive than using text prompts. Images convey more information and detail about styles than lengthy textual descriptions. A fixed text prompt generates high-quality results and assists in image generation. During the propagation process, edge detection maps of the captured camera images are used to supervise and ensure that while the appearance style is edited, the scene's shape is preserved.

The contributions of our work are mainly in three aspects: 1) The newly developed 3D style transfer pipeline enables 3D style transfer with zero samples. This pipeline allows for editing scene appearance styles by inputting style images directly, without requiring pre-training, while ensuring fast rendering and maintaining multi-view consistency. 2) We improve the IDU strategy, achieving faster training and inference speeds for pre-existing 3D scene edits. 3) Extensive experiments demonstrate that our method achieves superior 3D style transfer with high-quality stylization, offering improvements in both speed and performance compared to previous 3D editing methods.

\section{Related Works}
Currently, there are several approaches to 3D model editing, each with distinct focuses. One category utilizes diffusion editing methods, which combine specific loss functions to extend 2D image editing based on text prompts to 3D, thus enabling 3D model editing. Another approach integrates large-language models (LLMs) to locate target areas and modify their color and texture.

\subsection{3D Representations}

Various 3D representations have been proposed to address corresponding tasks. The first and most groundbreaking work is Neural Radiance Field (NeRF) \cite{ref11}, which uses volume rendering to complete reconstruction tasks with only 2D image supervision, fully preserving the information in the scene. However, due to this comprehensive data preservation, NeRF can be very time-consuming in reconstruction \cite{ref12,ref13,ref14} and subsequent editing \cite{ref15,ref16}.

Recently, 3D Gaussian Splatting \cite{ref7} has gradually been replacing NeRF. It has demonstrated impressive quality and speed in reconstruction tasks and has been widely cited in the generation field \cite{ref17,ref18}. Its efficient differentiable rendering and model design reduce space sampling in the scene, enabling faster training.
In this work, we apply 3D Gaussian Splatting (GS) to 3D scene texture stylization editing tasks to achieve fast and controllable editing.

\subsection{Generating 3D Content}
SPIn-NeRF\cite{ref19} focuses on object removal by initializing the multi-view mask of the target object. It uses the RGB image of the removed object as the appearance prior and the depth image as the geometric prior, performing joint optimization. Nevertheless, the resulting texture may only sometimes be consistent with the original scene. RePaint-NeRF \cite{ref20} also edits objects through masks. Utilizing masks and text prompts, it generates new content under the guidance of pre-trained diffusion and CLIP models, resulting in a multi-view consistent and scene-complete model. However, there are instances where the shape of the object changes significantly. Recent studies have applied the 2D text conditional diffusion model to 3D generation. DreamFusion \cite{ref21} proposed SDS loss as the loss function for NeRF-rendered images, which generates 3D models that are consistent at any angle. DreamGaussians \cite{ref22} applied the same concept as DreamFusion to GS, greatly speeding up the model generation process.

Our work is not to generate new scenes but to focus on editing to achieve more realistic editing and generation that aligns with the requirements of given GS scenes.

\begin{figure*}[htbp]
\centering
\includegraphics[width=5in]{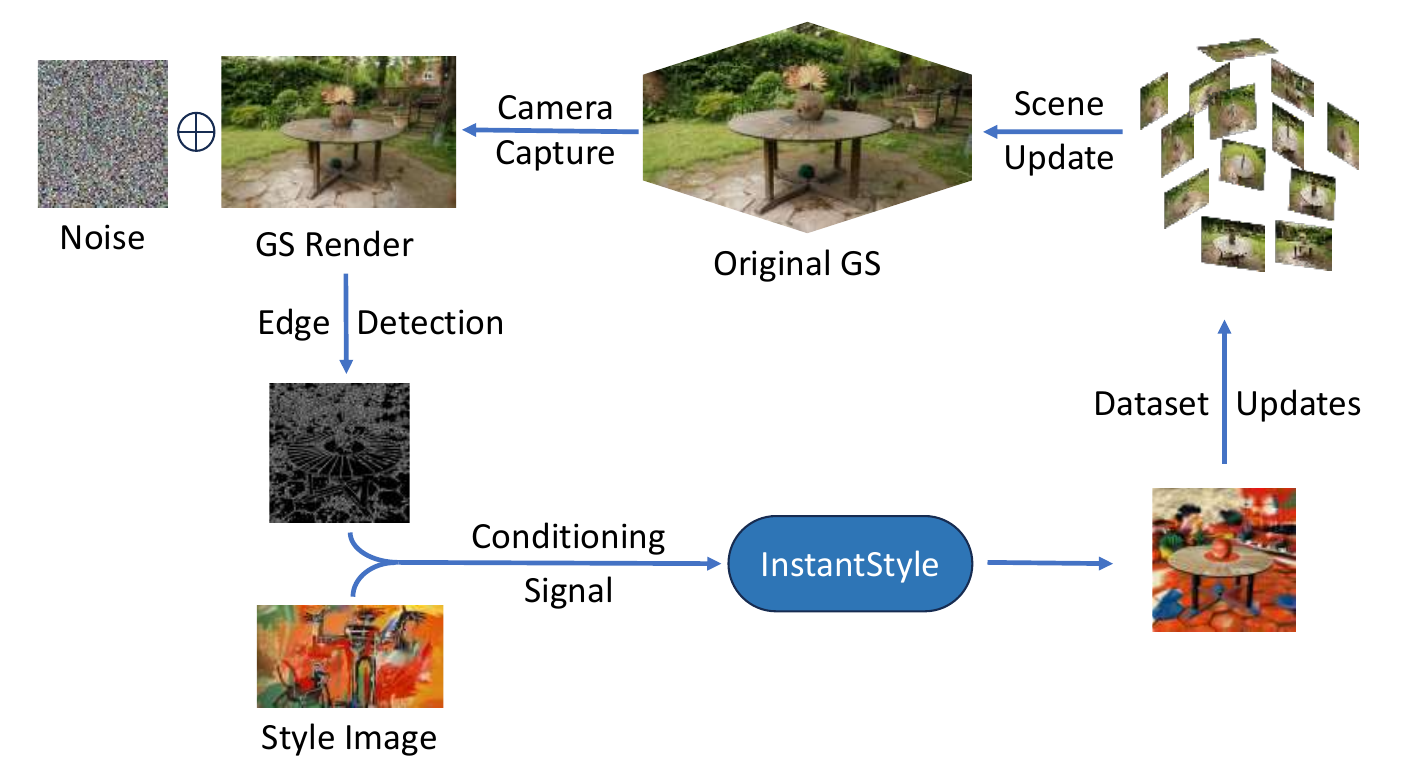}
\caption{Overview: Our method iteratively updates a subset of the GS dataset images to edit and reconstruct the GS scenes: (1) capture rendered images from the reconstructed scene, (2) process these images and the specified style image through InstantStyle to generate new images, (3) add the new images to the training dataset, and (4) continuously iterate to update and optimize the GS scenes.}
\label{fig_2}
\end{figure*}

\subsection{Artistic style transfer and editing}
Most stylization methods require textual cues and implement editing via CLIP and latent representations, as demonstrated in StylizedNeRF \cite{ref23} and StyleRF \cite{ref24}. These works may have limitations in selecting specific edit areas within the scene. InstructNeRF2NeRF \cite{ref25} uses InstructPix2pix as an editing reference and exploits an Iterative Dataset Update strategy to adjust the color, shape, and texture of the 3D model according to the text prompts without masks. These edits are applied to the dataset images of the original generated scene.

The recent GaussianEditor \cite{ref26,ref27} implements text editing in GS. The basic method is similar to Instruct-NeRF2NeRF, but it also combines the SAM method for segmentation to achieve selectability of the editing area. Gaussian Grouping \cite{ref28} also demonstrates an efficient segmentation method for subsequent repair and object removal tasks. StyleGaussian \cite{ref29} proposes an efficient feature rendering strategy capable of swiftly rendering high-dimensional VGG features \cite{ref8}. However, this approach necessitates prior learning of the style image and its integration into the 3DGS features.

\section{Method}
Our method uses the input style and text to jointly guide the generation of new scenes in the trained 3DGS scenes. We utilize the diffusion model (InstantStyle \cite{ref30}) for 2D image style transfer and improve the foundational strategy of Iterative Dataset Update in InstructNeRF2NeRF \cite{ref25}. We combine this strategy with the Nearest Neighbor Feature Matching (NNFM) loss, as proposed by ARF \cite{ref15}. The GS scenes are trained until they converge to the edit target while maintaining the consistency of the 3DGS scenes throughout the process.

\subsection{Background}
\textbf{3D Gaussian Splatting:}Gaussian Splatting \cite{ref7} is a state-of-the-art 3D scene representation that starts from a sparse point cloud to create an optimized 3D Gaussian point cloud. The 3D scene is represented as a set of 3D Gaussian primitives $g_p$= ($\mu _p$, $\sum_p$, $\sigma_p$, $c_p$) ,each Gaussian point characterized by a coordinate center mean $\mu_p\in R^{3}$, covariance matrix ${\textstyle \sum_{p}} \in R^{3\times 3} $, opacity $\sigma _{p}\in R^{3} $, and color $c_{p}\in R^{3} $.Compute the pixel color $C$ by mixing $N$ ordered Gaussians of overlapping pixels, effectively rendering a 3D Gaussian:
\begin{equation}
C=\sum_{i\in N}^{} c_{i} \alpha _{i}\prod_{j=1}^{i-1} (1-\alpha _{j} )
\end{equation}
where $c_i$ and $\alpha_i$ represent the color and density of the point, respectively, which are determined by the Gaussian covariance matrix and then scaled by the optimizable opacity and spherical harmonic color function. Through fast differentiable rasterization, 3DGS is faster to train than NeRF \cite{ref11}. The differentiable nature of the rasterizer allows edits made to the 2D image to propagate back, altering the underlying Gaussian scene data. Our method can quickly perform style editing on the scene based on the 3DGS representation.

\textbf{Style Transfer:}This visual stylization is time-consuming, from aligning the VGG features of the style image at the beginning to extending the style transfer to the 3D field \cite{ref26}. It has limited universality for untrained style images. This work employs the InstantStyle \cite{ref30} diffusion model to generate 2D images in various artistic styles based on an input style image and textual prompts. The results are then back-propagated to the 3DGS scene using specific loss functions. During the editing process, edge detection maps are additionally input to maintain the basic structure of the scene. Ultimately, the style editing of the entire scene is accomplished while preserving the original content.

\textbf{NNFM Loss:}	Artistic Radiance Fields (ARF) \cite{ref15} proposed a novel method to fuse 3D scenes and style images. The style image and scene rendering images are processed by a VGG encoder, and NNFM loss is applied to transfer complex high-frequency visual details from 2D style images to 3D scenes, better ensuring the preservation of local texture details. $F_s$ and $F_r$ are the scene renderings from the style image and the selected viewpoint, $F\left (i, j  \right )$ represents the feature vector at the feature map pixel position $\left (i, j  \right )$. The loss function can be written as:
\begin{equation}
L_{NNFM}\left (F_{r},F_{s}\right ) =\frac{1}{N} \sum_{i,j}\min_{i^{'},j^{'} } D\left ( F_{r}(i,j),  F_{s}(i^{'},j^{'})\right )
\end{equation}
where $N$ is the number of pixels in the $F_r$ rendered image, and $D\left (u_1,u_2\right )$ is the cosine distance between the two vectors $u_1$,$u_2$.

\subsection{InstantStyleGaussian}
Our method uses the given style image and text prompts for editing, primarily relying on image prompts, with text assisting in generating training datasets. The reconstructed scene model is fine-tuned to match the reference edited image style, ultimately generating a scene that conforms to the artistic style while maintaining 3D consistency, as shown in the Figure 2. The image-conditional diffusion model is used to iteratively update the training dataset images, which helps maintain the structural features of the original scene. In this section, we first describe the dataset image editing process based on the diffusion model, followed by the iterative process of optimizing the reconstruction of the 3DGS scene.

\textbf{Editing a rendered image:}We use the InstantStyle denoising diffusion model to update the dataset images, which accepts four inputs: the image to be updated $C_r$, textual prompts $C_t$, a style image $C_s$, and random noise input $Z_n$.The input image is captured from the camera's perspective, set as the conditional image, and transformed into latent variables to be fed into the U-Net network of the diffusion model. Then, based on the given textual prompts and style images, the model is trained by iteratively refining the noise. After decoding, it ultimately generates images that meet the specified criteria. In this way, the edited images are directly back-propagated to the GS scene, causing it to continuously update according to the input prompts, thereby generating new GS scenes that meet the conditions.In this process, edge operators are applied to the image to obtain the corresponding edge detection map before feeding the captured camera image into the U-Net network. This is then fed into the diffusion model along with the latent variables. This approach helps maintain the original structure in the generated image, ensuring that the shape and structure of the original objects do not undergo significant changes when transmitted to the GS scene.

\textbf{Iterative Dataset Update:}The InstructNeRF2NeRF method is an iterative process involving NeRF rendering images, editing images, and updating NeRF. While this approach is highly effective, it often requires significant training time due to the mixed training of edited and unedited images in the dataset. In this work, we propose a slight modification to this method. The core remains the Iterative Dataset Update process. However, we randomly select thirty or fewer camera perspectives (adjusted based on scene size) and perform a single edit on these images using a diffusion model. The edited images serve as references and increase the training dataset without replacing the original images from the corresponding perspectives. This continuous addition and optimization process enhances the GS scenes, ensuring improvement and refinement.However, the diffusion model may generate inconsistent, edited images from different perspectives. Unlike NeRF, 3DGS does not require training with light rays and can directly start from rendered images. Therefore, we use L1 and LPIPS \cite{ref34} losses to train Gaussian Splatting. For locally inconsistent textures from different perspectives, we used NNFM Loss to match local features, thereby better-preserving texture details.

\subsection{Implementation details}
We use the gsplat library from GaussianEditor \cite{ref27} as the underlying model and visualization tool, with InstantStyle as the diffusion model. Several additional parameters, such as the ControlNet conditioning scale and the guiding weights for text and image adjustments, determine the diffusion model's updated intensity. These parameters are set to default values, but some hyperparameters have been modified to adapt to the Gaussian distribution. The relevant guiding weights can be manually adjusted to achieve the desired editing effect in scenarios where the generated results are suboptimal.

Our training method involves a maximum of 1k iterations. On an A100 GPU (40GB of memory), it takes only twenty minutes to complete the style transfer editing of a scene. The first fifteen minutes are dedicated to the diffusion model generating images, and these images are backpropagated through the entire scene in just five minutes. The final editing result is a subjective judgment; longer training times may lead to extreme editing effects, so training is stopped once convergence is achieved.

\section{Results}
\textbf{Datasets and Baselines.} We evaluate our method on the Tanks and Temples datasets \cite{ref35} and the Mip-NeRF 360 datasets \cite{ref12}. Each dataset ranges in size from 150 to 400 images. These datasets feature complex geometries and realistic real-world textures, making the generation of new scenes more meaningful and realistic.

\begin{figure*}[htbp]
\centering
\includegraphics[width=6in]{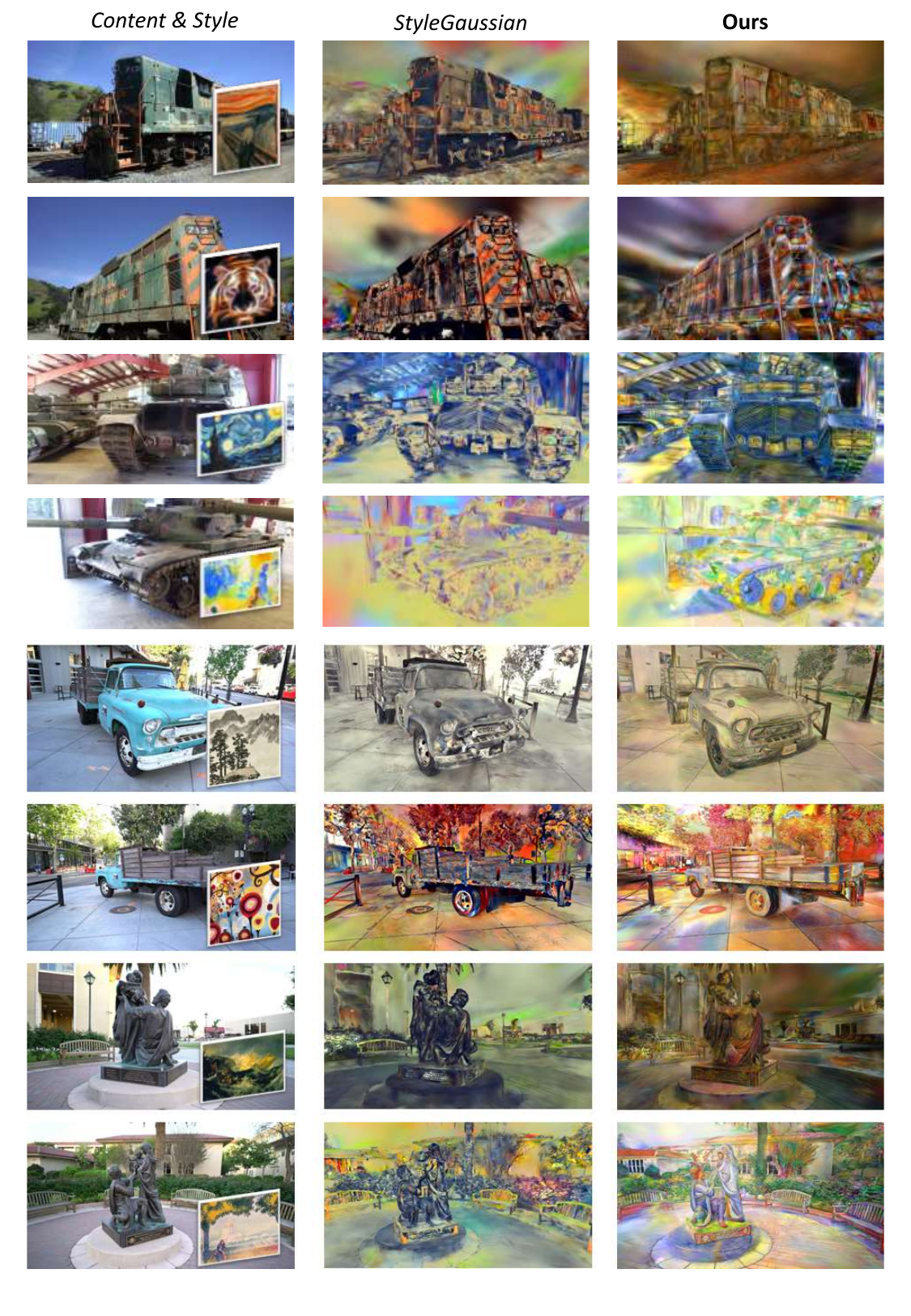}
\caption{Qualitative Evaluation. Compared to the effects of StyleGaussian \cite{ref29}, our method demonstrates superior style transfer quality, better matching the reference style while preserving the original content more effectively.}
\label{fig_3}
\end{figure*}

\subsection{Qualitative Evaluation}
We show the qualitative results in Figure 3. InstantStyleGaussian demonstrates exceptional style transfer performance, generating high-quality 3D scenes with better artistic style alignment to the reference images and improved preservation of the original scene content. StyleGaussian uses a decoded stylized RGB design, and the quality of the generated scenes depends on the decoded features. For the styles in rows 2, 4, and 8 in Figure 3, it produced colors that did not match the reference style images. Our method employs a diffusion model that captures style features more effectively without additional training, and the input edge detection images help preserve the original scene's textures and content.

\subsection{Ablations}
\textbf{Nearest Neighbor Feature Matching (NNFM).} The introduced NNFM Loss aids in better handling local detail textures and fitting inconsistent colors. As shown in the third and fourth images of each row in Figure 4, without NNFM Loss, the rendered results exhibit significantly darker overall colors and a loss of control over color textures in local regions.

\begin{figure}[htbp]
\centering
\includegraphics[width=3.5in]{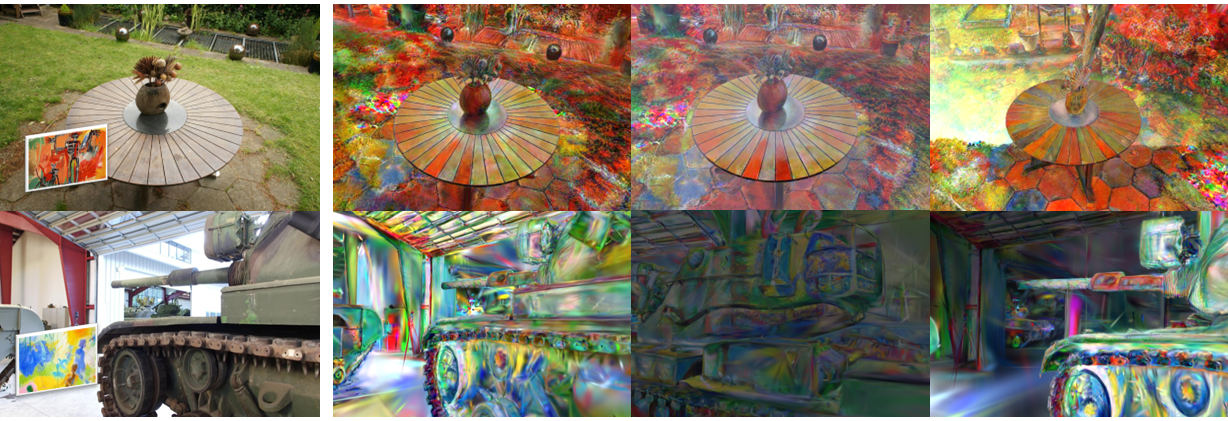}
\caption{Without NNFM Loss, the quality of style transfer significantly decreases and does not maintain multi-view consistency.}
\label{fig_4}
\end{figure}

\textbf{More iterations.} The final style transfer effect is subjective, but merely increasing the number of iterations does not improve the overall scene quality. Instead, it can lead to overfitting in local regions, causing textures to become abnormal, as indicated by the red boxes in Figure 5.

\begin{figure}[htbp]
\centering
\includegraphics[width=3.5in]{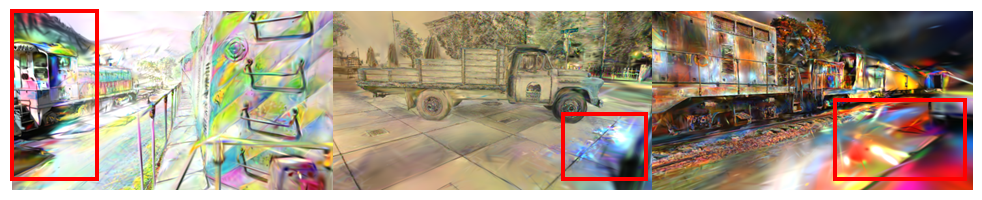}
\caption{Increasing the number of iterations leads to overfitting of textures in local regions.}
\label{fig_5}
\end{figure}

\subsection{Quantitative Evaluation}
As previous works \cite{ref15,ref17} pointed out, a standardized metric currently needs to evaluate the quality of 3D style transfer. To quantitatively compare multi-view consistency and rendering time, we follow the approach from prior work: We warp one view to another according to the optical flow \cite{ref37} using softmax splatting \cite{ref38} and then compute the masked RMSE score and LPIPS score \cite{ref34} to measure stylization consistency. Our method ensures scene quality while outperforming previous works in performance. The results are summarized in Table 1.

Unlike our method, StyleGaussian requires Feature Embedding Training and Style Transfer Training. During Style Transfer Training, it needs to learn styles from images in a predefined dataset such as WikiArt\cite{ref41}, which requires substantial training time. Once trained, the model can quickly perform style transfers. In contrast, our method can directly infer and update 3D scenes using the input style image as a reference without requiring additional training. Our method significantly reduces the time required for the style transfer process while optimizing the transfer procedure. It does not need to repeatedly regenerate novel views for each perspective, resulting in a substantial speed improvement and a 13-fold increase in overall editing speed.

\renewcommand{\arraystretch}{1.1}
\begin{table}[h]
\caption{Quantitative results}
\centering
\scalebox{0.95}{
\begin{tabular}{c|c c|c c|c}
\hline
\multirow{2}{*}[0.5em]{Methods} & \multicolumn{2}{c|}{\makecell{Short-range \\ Consistency}} & \multicolumn{2}{c|}{\makecell{Long-range \\ Consistency}} & \multirow{2}{*}[0.6em]{\makecell{Total \\ Editing Time}} \\
\hline
 & LPIPS & RMSE & LPIPS & RMSE &  \\
\textit{StyleGaussian\cite{ref29}} & 0.028 & 0.045 & 0.082 & 0.085 & 270 minutes \\
\textbf{Ours} & \textbf{0.024} & \textbf{0.026} & \textbf{0.074} & \textbf{0.076} & \textbf{20 minutes} \\
\hline
\end{tabular}
}
    \begin{tablenotes}
		\footnotesize
		\item We use LPIPS (↓) and RMSE (↓) to evaluate the consistency performance of InstantStyleGaussian compared to state-of-the-art methods. We also assess the time required to complete the editing of an entire scene.
    \end{tablenotes}
\end{table}

\section{Limitations}
Our method primarily focuses on texture editing of scene surfaces, making other edits ineffective. For instance, removing existing objects from the scene through segmentation can result in significant artifacts due to the need for original information about where the objects were located. Similarly, adding new objects to the scene is challenging as placing them accurately in the specified positions is difficult. However, we believe that adding and removing content will also become feasible with the advent of more powerful diffusion models for image processing. Additionally, InstantStyleGaussian modifies the color properties in the Gaussian function while keeping the geometry unchanged, making it unsuitable for style transfer involving geometric deformations. Incorporating geometric deformation into style transfer is a potential research direction that may require more effective supervision.

\section{Conclusion}
This paper introduced InstantStyleGaussian, an innovative 3D style transfer method that can quickly generate new 3D GS scenes by specifying a target style image. Our approach operates on pre-reconstructed GS scenes, combining diffusion models with an improved Iterative Dataset Update strategy. This ensures fast style transfer while maintaining fast rendering and multi-view consistency. We demonstrated high-quality results across various scenes, and our method has potential applications in game development, virtual reality, and augmented reality.

\section{Acknowledgment}
This research was supported by the Baima Lake Laboratory Joint Funds of the Zhejiang Provincial Natural Science Foundation of China under Grant No. LBMHD24F030002 and the National Natural Science Foundation of China under Grant 62373329.

\begin{figure*}[htbp]
\appendix
\centering
\includegraphics[width=7.2in]{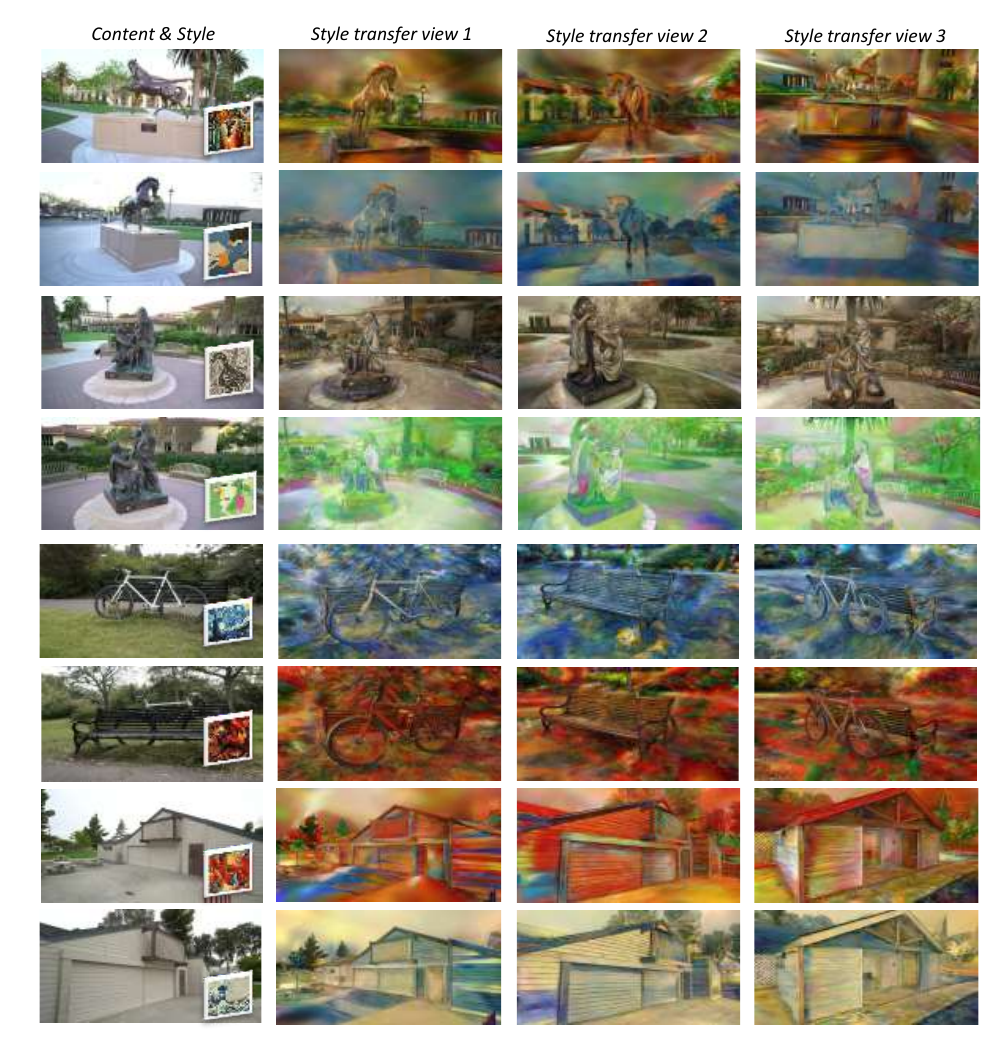}
\caption{Additional stylization results in more scenes. These examples further demonstrate the versatility and effectiveness of our method in various environments.}
\label{fig_6}
\end{figure*}


\begin{thebibliography}{00}
\bibitem{ref1}
Haque, A., Tancik, M., Efros, A. A., Holynski, A., $\&$ Kanazawa, A. (2023). Instruct-nerf2nerf: Editing 3d scenes with instructions. In Proceedings of the IEEE/CVF International Conference on Computer Vision (pp. 19740-19750).

\bibitem{ref2}
Park, J., Kwon, G., $\&$ Ye, J. C. (2023). ED-NeRF: Efficient Text-Guided Editing of 3D Scene using Latent Space NeRF. arxiv preprint arxiv:2310.02712.

\bibitem{ref3}
Wang, C., Chai, M., He, M., Chen, D., $\&$ Liao, J. (2022). Clip-nerf: Text-and-image driven manipulation of neural radiance fields. In Proceedings of the IEEE/CVF Conference on Computer Vision and Pattern Recognition (pp. 3835-3844).

\bibitem{ref4}
Wang, C., Jiang, R., Chai, M., He, M., Chen, D., $\&$ Liao, J. (2023). Nerf-art: Text-driven neural radiance fields stylization. IEEE Transactions on Visualization and Computer Graphics.

\bibitem{ref5}
Garbin, S. J., Kowalski, M., Johnson, M., Shotton, J., $\&$ Valentin, J. (2021). Fastnerf: High-fidelity neural rendering at 200fps. In Proceedings of the IEEE/CVF international conference on computer vision (pp. 14346-14355).

\bibitem{ref6}
Huang, X., Li, W., Hu, J., Chen, H., $\&$ Wang, Y. (2023). Refsr-nerf: Towards high fidelity and super resolution view synthesis. In Proceedings of the IEEE/CVF Conference on Computer Vision and Pattern Recognition (pp. 8244-8253).

\bibitem{ref7}
Kerbl, B., Kopanas, G., Leimkühler, T., $\&$ Drettakis, G. (2023). 3D Gaussian Splatting for Real-Time Radiance Field Rendering. ACM Trans. Graph., 42(4), 139-1.

\bibitem{ref8}
Simonyan, K. (2014). Very deep convolutional networks for large-scale image recognition. arxiv preprint arxiv:1409.1556.

\bibitem{ref9}
Huang, H. P., Tseng, H. Y., Saini, S., Singh, M., $\&$ Yang, M. H. (2021). Learning to stylize novel views. In Proceedings of the IEEE/CVF International Conference on Computer Vision (pp. 13869-13878).

\bibitem{ref10}
Gu, J., Liu, L., Wang, P., $\&$ Theobalt, C. (2021). Stylenerf: A style-based 3d-aware generator for high-resolution image synthesis. arxiv preprint arxiv:2110.08985.

\bibitem{ref11}
Mildenhall, B., Srinivasan, P. P., Tancik, M., Barron, J. T., Ramamoorthi, R., $\&$ Ng, R. (2021). Nerf: Representing scenes as neural radiance fields for view synthesis. Communications of the ACM, 65(1), 99-106.

\bibitem{ref12}
Barron, J. T., Mildenhall, B., Verbin, D., Srinivasan, P. P., $\&$ Hedman, P. (2022). Mip-nerf 360: Unbounded anti-aliased neural radiance fields. In Proceedings of the IEEE/CVF conference on computer vision and pattern recognition (pp. 5470-5479).

\bibitem{ref13}
Hedman, P., Srinivasan, P. P., Mildenhall, B., Barron, J. T., $\&$ Debevec, P. (2021). Baking neural radiance fields for real-time view synthesis. In Proceedings of the IEEE/CVF international conference on computer vision (pp. 5875-5884).

\bibitem{ref14}
Müller, T., Evans, A., Schied, C., $\&$ Keller, A. (2022). Instant neural graphics primitives with a multiresolution hash encoding. ACM transactions on graphics (TOG), 41(4), 1-15.

\bibitem{ref15}
Zhang, K., Kolkin, N., Bi, S., Luan, F., Xu, Z., Shechtman, E., $\&$ Snavely, N. (2022, October). Arf: Artistic radiance fields. In European Conference on Computer Vision (pp. 717-733). Cham: Springer Nature Switzerland.

\bibitem{ref16}
Müller, T., Evans, A., Schied, C., $\&$ Keller, A. (2022). Instant neural graphics primitives with a multiresolution hash encoding. ACM transactions on graphics (TOG), 41(4), 1-15.

\bibitem{ref17}
Chen, Z., Wang, F., Wang, Y., $\&$ Liu, H. (2024). Text-to-3d using gaussian splatting. In Proceedings of the IEEE/CVF Conference on Computer Vision and Pattern Recognition (pp. 21401-21412).

\bibitem{ref18}
Yi, T., Fang, J., Wu, G., Xie, L., Zhang, X., Liu, W., ... $\&$ Wang, X. (2023). Gaussiandreamer: Fast generation from text to 3d gaussian splatting with point cloud priors. arxiv preprint arxiv:2310.08529.

\bibitem{ref19}
Mirzaei, A., Aumentado-Armstrong, T., Derpanis, K. G., Kelly, J., Brubaker, M. A., Gilitschenski, I., $\&$ Levinshtein, A. (2023). SPIn-NeRF: Multiview segmentation and perceptual inpainting with neural radiance fields. In Proceedings of the IEEE/CVF Conference on Computer Vision and Pattern Recognition (pp. 20669-20679).

\bibitem{ref20}
Zhou, X., He, Y., Yu, F. R., Li, J., $\&$ Li, Y. (2023). Repaint-nerf: Nerf editting via semantic masks and diffusion models. arxiv preprint arxiv:2306.05668.

\bibitem{ref21}
Poole, B., Jain, A., Barron, J. T., $\&$ Mildenhall, B. (2022). Dreamfusion: Text-to-3d using 2d diffusion. arxiv preprint arxiv:2209.14988.

\bibitem{ref22}
Tang, J., Ren, J., Zhou, H., Liu, Z., $\&$ Zeng, G. (2023). Dreamgaussian: Generative gaussian splatting for efficient 3d content creation. arxiv preprint arxiv:2309.16653.

\bibitem{ref23}
Huang, Y. H., He, Y., Yuan, Y. J., Lai, Y. K., $\&$ Gao, L. (2022). Stylizednerf: consistent 3d scene stylization as stylized nerf via 2d-3d mutual learning. In Proceedings of the IEEE/CVF Conference on Computer Vision and Pattern Recognition (pp. 18342-18352).

\bibitem{ref24}
Liu, K., Zhan, F., Chen, Y., Zhang, J., Yu, Y., El Saddik, A., ... $\&$ Xing, E. P. (2023). Stylerf: Zero-shot 3d style transfer of neural radiance fields. In Proceedings of the IEEE/CVF Conference on Computer Vision and Pattern Recognition (pp. 8338-8348).

\bibitem{ref25}
Brooks, T., Holynski, A., $\&$ Efros, A. A. (2023). Instructpix2pix: Learning to follow image editing instructions. In Proceedings of the IEEE/CVF Conference on Computer Vision and Pattern Recognition (pp. 18392-18402).

\bibitem{ref26}
Wang, J., Fang, J., Zhang, X., Xie, L., $\&$ Tian, Q. (2024). Gaussianeditor: Editing 3d gaussians delicately with text instructions. In Proceedings of the IEEE/CVF Conference on Computer Vision and Pattern Recognition (pp. 20902-20911).

\bibitem{ref27}
Chen, Y., Chen, Z., Zhang, C., Wang, F., Yang, X., Wang, Y., ... $\&$ Lin, G. (2024). Gaussianeditor: Swift and controllable 3d editing with gaussian splatting. In Proceedings of the IEEE/CVF Conference on Computer Vision and Pattern Recognition (pp. 21476-21485).

\bibitem{ref28}
Ye, M., Danelljan, M., Yu, F., $\&$ Ke, L. (2023). Gaussian grouping: Segment and edit anything in 3d scenes. arXiv preprint arXiv:2312.00732.

\bibitem{ref29}
Liu, K., Zhan, F., Xu, M., Theobalt, C., Shao, L., $\&$ Lu, S. (2024). StyleGaussian: Instant 3D Style Transfer with Gaussian Splatting. arXiv preprint arXiv:2403.07807.

\bibitem{ref30}
Wang, H., Wang, Q., Bai, X., Qin, Z.,$\&$ Chen, A. (2024). Instantstyle: Free lunch towards style-preserving in text-to-image generation. arXiv preprint arXiv:2404.02733.

\bibitem{ref34}
Zhang, R., Isola, P., Efros, A. A., Shechtman, E., $\&$ Wang, O. (2018). The unreasonable effectiveness of deep features as a perceptual metric. In Proceedings of the IEEE conference on computer vision and pattern recognition (pp. 586-595).

\bibitem{ref35}
Knapitsch, A., Park, J., Zhou, Q. Y., $\&$ Koltun, V. (2017). Tanks and temples: Benchmarking large-scale scene reconstruction. ACM Transactions on Graphics (ToG), 36(4), 1-13.

\bibitem{ref37}
Teed, Z., $\&$ Deng, J. (2020). Raft: Recurrent all-pairs field transforms for optical flow. In Computer Vision–ECCV 2020: 16th European Conference, Glasgow, UK, August 23–28, 2020, Proceedings, Part II 16 (pp. 402-419). Springer International Publishing.

\bibitem{ref38}
Niklaus, S., $\&$ Liu, F. (2020). Softmax splatting for video frame interpolation. In Proceedings of the IEEE/CVF conference on computer vision and pattern recognition (pp. 5437-5446).

\bibitem{ref39}
WikiArt, W. (2018). Visual Art Encyclopedia.

\bibitem{ref40}
Jaganathan, V., Huang, H. H., Irshad, M. Z., Jampani, V., Raj, A., $\&$ Kira, Z. (2024). ICE-G: Image Conditional Editing of 3D Gaussian Splats. arXiv preprint arXiv:2406.08488.

\bibitem{ref41}
Chiang, P. Z., Tsai, M. S., Tseng, H. Y., Lai, W. S., $\&$ Chiu, W. C. (2022). Stylizing 3d scene via implicit representation and hypernetwork. In Proceedings of the IEEE/CVF Winter Conference on Applications of Computer Vision (pp. 1475-1484).

\end{thebibliography}
\end{document}